\relax
\documentclass{article}
\usepackage{ijcai19}

\usepackage{times}  
\usepackage{helvet}  
\usepackage{courier}  
\usepackage{url}  
\usepackage{graphicx}  
\usepackage[utf8]{inputenc}
\usepackage{footnote}
\usepackage{tablefootnote}
\usepackage{comment}
\usepackage[para,online,flushleft]{threeparttable}

\usepackage{amsfonts,bm,color}
\usepackage{booktabs}       
\usepackage{amsopn}
\usepackage{amsmath}
\usepackage{amssymb}
\usepackage{empheq}
\usepackage{epstopdf}
\usepackage{float}
\usepackage{framed}
\usepackage{multirow}

\usepackage{adjustbox}
\usepackage{algorithm,algorithmicx}
\usepackage{algpseudocode}
\usepackage[small]{caption}

\usepackage{subcaption}

\newtheorem{myprop}{\bf{Proposition}}

\newcommand{\argmin}{\operatornamewithlimits{arg\,min}}

\DeclareMathOperator*{\minimize}{\text{minimize}}
\DeclareMathOperator*{\maximize}{\text{maximize}}
\DeclareMathOperator*{\st}{\text{subject to}}
\DeclareMathAlphabet\mathbfcal{OMS}{cmsy}{b}{n}
\newcommand{\Def}[0]{\mathrel{\mathop:}=}

\definecolor{Sijia_color}{rgb}{0.858, 0.188, 0.478}
\definecolor{Hongge_color}{rgb}{0.188, 0.858, 0.478}

\definecolor{aquamarine}{rgb}{0.5, 1.0, 0.83}
\definecolor{orange-red}{rgb}{1.0, 0.27, 0.0}
\definecolor{gray}{rgb}{0.5, 0.5, 0.5}
\definecolor{blue}{rgb}{0.0, 0.0, 1.0}
\definecolor{fuchsia}{rgb}{1.0, 0.0, 1.0}
\definecolor{black}{rgb}{0.0, 0.0, 0.0}
\begin{document}

\title{Topology Attack and Defense for Graph Neural Networks:\\An Optimization Perspective}

\author{
Kaidi Xu$^1$\footnote{Equal contribution}\and
Hongge Chen$^{2*}$\and
Sijia Liu$^{3}$\and
Pin-Yu Chen$^3$ \and 
Tsui-Wei Weng$^2$\and \\
Mingyi Hong$^4$\And 
Xue Lin$^1$\\
\affiliations
$^1$Electrical \& Computer Engineering, Northeastern University, Boston, USA\\
$^2$Electrical Engineering \& Computer Science, Massachusetts Institute of Technology, Cambridge, USA \\
$^3$MIT-IBM Watson AI Lab, IBM Research\\
$^4$Electrical \& Computer Engineering, University of Minnesota, Minneapolis, USA\\
\emails
xu.kaid@husky.neu.edu, 
chenhg@mit.edu, 
sijia.Liu@ibm.com, 
pin-yu.chen@ibm.com, 
twweng@mit.edu, 
mhong@umn.edu, 
xue.lin@northeastern.edu
}

\maketitle

\begin{abstract}
Graph neural networks (GNNs) which apply the deep neural networks to graph data have achieved significant performance for the task of semi-supervised node classification. However, only few work has addressed the adversarial robustness of GNNs. In this paper, we first present a novel gradient-based attack method that facilitates the difficulty of tackling discrete graph data.
When comparing to current adversarial attacks on GNNs, the results show that by only perturbing a small number of edge perturbations, including addition and deletion, our optimization-based attack can lead to a noticeable decrease in classification performance. Moreover, leveraging our gradient-based attack, we propose the first optimization-based adversarial training for GNNs. Our method yields higher robustness against both different gradient based and greedy attack methods without sacrificing classification accuracy on original graph. Code is available at \url{https://github.com/KaidiXu/GCN_ADV_Train}.
\end{abstract}

\section{Introduction}
\label{sec: intro}
Graph structured data plays a crucial role in many AI applications. It is an important and versatile representation to model a wide variety of datasets from many domains, such as
molecules, social networks, or interlinked documents with citations. Graph neural networks (GNNs) on  graph structured data have shown outstanding results in various applications \cite{kipf2016semi,velivckovic2017graph,xu2018powerful}. However, despite the great success on inferring from graph data, the inherent 
challenge of lacking adversarial robustness in deep learning models still carries over to security-related domains such as blockchain or communication networks.

In this paper, we aim to evaluate the robustness of GNNs  from a perspective of first-order optimization adversarial attacks.
It is worth mentioning that first-order methods have achieved  great success for generating adversarial attacks on audios or images \cite{carlini2018audio,xu2019interpreting,chen2017ead,xu2018structured,chen2018attacking}. However, some recent works \cite{dai2018adversarial,bojcheski2018adversarial} suggested that conventional (first-order) continuous optimization methods do not directly apply to attacks using edge manipulations (we call \textit{topology attack}) due to the discrete nature of graphs.
We close this gap by studying the problem of generating topology attacks via convex relaxation
so that gradient-based adversarial attacks become plausible for GNNs. Benchmarking on node classification tasks using GNNs, our gradient-based topology attacks     outperform  current state-of-the-art attacks  subject to the same topology perturbation budget. This demonstrates the effectiveness of our attack generation method through the lens of convex relaxation and first-order optimization. Moreover, by leveraging our proposed gradient-based attack, we propose the first optimization-based adversarial training technique for GNNs, yielding significantly improved robustness against gradient-based and greedy topology attacks.


Our new attack generation and adversarial training methods for GNNs are built upon the theoretical foundation of spectral graph theory, first-order optimization,  and robust (mini-max) optimization.
We summarize our main contributions as follows:
\begin{itemize}
    \item We propose a general first-order  attack generation framework  under  two attacking
   scenarios: a) attacking a pre-defined GNN  and  b) attacking a re-trainable  GNN. This yields two new topology attacks:  projected gradient descent (PGD) topology attack and min-max topology attack.
    Experimental results 
    show that 
    the proposed attacks   outperform  current state-of-the-art attacks.
    \item With the aid of our first-order attack generation methods, we propose an adversarial training method for GNNs to improve their robustness. The effectiveness of our method is shown by the considerable improvement of robustness  on GNNs   against both optimization-based  and greedy-search-based topology attacks. 
\end{itemize} 
\section{Related Works}
 Some recent attentions have been paid to the robustness of graph neural network. Both \cite{zugner2018adversarial} and \cite{dai2018adversarial} studied adversarial attacks on neural networks for graph data. \cite{dai2018adversarial} studied test-time non-targeted adversarial attacks on both graph classification and node classification. Their work restricted the attacks to perform modifications on discrete structures, that is, an attacker is only allowed to add or delete edges from a graph to construct a new graph. White-box, practical black-box and restricted black-box graph adversarial attack scenarios were studied. Authors in \cite{zugner2018adversarial} considered both test-time (evasion) and training-time (data poisoning) attacks on node classification task. In contrast to~\cite{dai2018adversarial}, besides adding or removing edges in the graph, attackers in~\cite{zugner2018adversarial} may modify node attributes. They designed adversarial attacks based on a static surrogate model and evaluated their impact
by training a classifier on the data modified by the attack. The resulting attack algorithm is for targeted attacks on single nodes. It was shown that small perturbations on the graph structure and
node features are able to achieve misclassification of a target node. 
 A data poisoning attack on unsupervised node representation learning, or node embeddings, has been proposed in~\cite{bojcheski2018adversarial}. This attack is based on perturbation theory to maximize the loss obtained from DeepWalk~\cite{perozzi2014deepwalk}.
 In \cite{zugner2019adversarial},  training-time attacks on GNNs were also  investigated for node classification by perturbing
the graph structure.
The authors solved
a min-max problem in training-time attacks using meta-gradients and treated the graph topology
as a hyper-parameter to optimize.

\section{Problem Statement}
\label{sec: GNN}

We begin by providing  preliminaries  on GNNs. We then formalize the attack threat  model of GNNs in terms of  edge perturbations, which we refer as `topology attack'.  

\subsection{Preliminaries on GNNs} 
It has been recently shown in \cite{kipf2016semi,velivckovic2017graph,xu2018powerful} that GNN is powerful in transductive learning, e.g., node classification under graph data. That is, given a single network topology with node features and a known subset of node labels, GNNs are efficient to infer the classes of unlabeled nodes. Prior to defining GNN, we first introduce the following graph notations.  Let $\mathcal G = (\mathcal V, \mathcal E)$ denote an undirected and unweighted graph, where $\mathcal V$ is the vertex (or node) set with cardinality $|\mathcal V| = N$, and $\mathcal E \in (\mathcal V \times \mathcal V)$ denotes the edge set with cardinality $|\mathcal E| = M$. Let $\mathbf  A$ represent a binary adjacency  matrix. 
By definition, we have $A_{ij} = 0$ if $(i, j) \notin \mathcal E$. In a GNN, 
we assume that each node $i$ is associated with a feature vector  $\mathbf x_i \in \mathbb R^{M_0}$ and a scalar label $y_i$. The goal of GNN is to predict the class of an unlabeled node under the graph topology $\mathbf A$ and the training data $\{ (\mathbf x_i, y_i )\}_{i=1}^{N_{\text{train}}}$. Here GNN uses  input features of all nodes but only 
$N_{\text{train}} < N$  nodes with labeled classes in the training phase.


Formally, the 
$k$th layer of a GNN model
obeys the propagation rule of the generic form 
{\small \begin{align}\label{eq:  activation_GNN}
    \mathbf h_i^{(k)} = g^{(k)} \left ( \{ \mathbf W^{(k-1)} \mathbf h_j^{(k-1)} \tilde A_{ij}, ~ \forall j \in \mathcal N(i)\} \right ), ~ \forall i\in [N]
\end{align}}%
where $\mathbf h_i^{(k)} \in \mathbb R^{M_{k}}$ denotes the feature vector of node $i$ at layer $k$,   
$\mathbf h_i^{(0)} = \mathbf x_i\in \mathbb R^{M_0}$ is the input feature vector of node $i$,
$ g^{(k)}$ is a possible composite mapping (activation) function, $\mathbf W^{(k-1)} \in \mathbb R^{M_k \times M_{k-1}}$ is  the trainable weight matrix at layer $(k-1)$, $\tilde { A}_{ij}$ is the $(i,j)$th entry of $\tilde {\mathbf A}$ that
 denotes a linear mapping of $\mathbf A$ but with the same sparsity pattern,
and $\mathcal N(i)$ denotes node $i$'s neighbors together with itself, i.e., $\mathcal N(i) = \{ j | (i,j) \in \mathcal E, \text{ or } j = i \}$. 

A special form of GNN is graph convolutional networks (GCN)~\cite{kipf2016semi}. This is a recent
approach of learning on graph structures using convolution operations which is promising as an embedding methodology.
In GCNs, the propagation rule  \eqref{eq: activation_GNN} becomes \cite{kipf2016semi}
{\small \begin{align}
        \mathbf h_i^{(k)} = \sigma \left (  \sum_{j \in \mathcal N_i} \left  ( \mathbf W^{(k-1)}  \mathbf h_j^{(k-1)} \tilde A_{ij} \right ) \right ),
\end{align}}%
where $\sigma (\cdot )$ is the ReLU function. 
Let $\tilde A_{i,:}$ denote  the $i$th row  of $\tilde {\mathbf A}$ and $\mathbf H^{(k)} = \left [ (\mathbf h_1^{(k)})^\top; \ldots; (\mathbf h_N^{(k)})^\top   \right ]$, we then have
the standard form of  GCN,
{\small \begin{align}
        \mathbf H^{(k)} = \sigma \left (  \tilde{\mathbf A} \mathbf H^{(k-1)} ( \mathbf W^{(k-1)} )^\top
        \right ).
\end{align}}%
Here $\tilde {\mathbf A}$ is given by a normalized adjacency matrix $\tilde{\mathbf A} = \hat{\mathbf D}^{-1/2} \hat{\mathbf A} \hat{\mathbf D}^{-1/2}$, where $\hat{\mathbf A} = \mathbf A + \mathbf I$, and $\hat {\mathbf D}_{ij}  = 0$ if $i \neq j$ and $\hat {\mathbf D}_{ii} = \mathbf 1^\top \hat{\mathbf{A}}_{:,i}$.


\subsection{Topology Attack in Terms of Edge Perturbation}
We introduce a Boolean symmetric matrix $\mathbf S \in \{ 0,1\}^{N \times N}$ to encode whether or not an edge in $\mathcal G$ is modified. That is,  the edge connecting nodes $i$ and $j$ is modified (added or removed) if and only if $S_{ij} = S_{ji}=1$. Otherwise, $S_{ij } = 0$ if $i = j $ or the edge $(i,j)$ is not perturbed.
Given the adjacency matrix $\mathbf A$, its supplement is given by $\bar {\mathbf A} = \mathbf 1 \mathbf 1^T - \mathbf I -   \mathbf A $, where $\mathbf I$ is an identity matrix, and 
$(\mathbf 1 \mathbf 1^T - \mathbf I)$ corresponds to the fully-connected graph. With the aid of edge perturbation matrix $\mathbf S$ and $\bar {\mathbf A}$, a perturbed graph topology $\mathbf A^\prime  $ against $\mathbf A$ is given by
{\small \begin{align}\label{eq:A_indci} 
    \mathbf A^\prime = \mathbf A + \mathbf C \circ \mathbf S, ~\mathbf C = \bar {\mathbf A} -   \mathbf A,
\end{align}}%
where  $\circ $ denotes the element-wise product. In \eqref{eq:A_indci}, the positive entry of $\mathbf C$ denotes the edge that can be added to the graph $\mathbf A$, and the negative entry of $\mathbf C$ denotes the edge that can be removed from  $\mathbf A$. 
We then formalize the concept of \textit{topology attack} to GNNs: Finding minimum edge perturbations encoded by $\mathbf S$ in \eqref{eq:A_indci} to mislead GNNs.  A more detailed attack formulation will be studied in the next section. 

\section{Topology Attack Generation: A First-Order Optimization Perspective}

In this section, we first define   attack loss  (beyond the conventional cross-entropy loss) under different attacking scenarios. We then develop two efficient  attack generation methods by leveraging first-order optimization. We  call the resulting attacks projected gradient descent (PGD) topology attack and min-max topology attack, respectively.

\subsection{Attack Loss \& Attack Generation}
Let $\mathbf Z (\mathbf S, \mathbf W; \mathbf A,  \{ \mathbf x_i\})$ denote the prediction probability of a GNN specified by $\mathbf A^\prime$ in \eqref{eq:A_indci} and $\mathbf W$ under input features $ \{ \mathbf x_i\}$. Then 
\textcolor{black}{$Z_{i,c}$}
denotes 
 the probability of assigning node $i$ to class $c$. It has been shown in existing works~\cite{goodfellow2014explaining,kurakin2016adversarial} that 
 the \textit{negative  cross-entropy (CE) loss} between the true labels ($ y_i$)   and the predicted labels ($\{ Z_{i,c} \}$) can be used as an attack loss at node   $i$, denoted by $f_i(\mathbf S, \mathbf W;\mathbf A,  \{ \mathbf x_i\}, y_i )$.   We can also propose a \textit{CW-type loss} similar to Carlili-Wagner (CW)  attacks for attacking image classifiers  \cite{carlini2017towards},
\textcolor{black}{{\small \begin{align}\label{eq:  CW_loss_nodei}
& f_i(\mathbf S, \mathbf W;  \mathbf A,  \{ \mathbf x_i\}, y_i )  
=
\max \left  \{   Z_{i,y_i} - \max_{c \neq y_i}  Z_{i,c}, - \kappa  \right \},
\end{align}}%
where $\kappa \geq  0$  is a confidence level of making wrong decisions.}


To design topology attack, we seek $\mathbf S$ in \eqref{eq:A_indci}  to minimize  the per-node attack loss (CE-type or CW-type) given  a finite budge of edge perturbations. We consider two threat models: a) attacking a pre-defined GNN  with known $\mathbf W$; b) attacking an interactive GNN with re-trainable $\mathbf W$. In the case a) of fixed $\mathbf W$, 
the attack generation problem can be cast as  
{\small \begin{align}\label{eq:  prob_attack_s}
    \begin{array}{ll}
\displaystyle \minimize_{\mathbf s }         &  \sum_{i \in \mathcal V}f_i(\mathbf s  ; \mathbf W,  \mathbf A,  \{ \mathbf x_i\}, y_i )\\
      \st    & \mathbf 1^\top \mathbf s \leq  \epsilon, ~ \mathbf s\in \{ 0,1\}^{n},
    \end{array}
\end{align}}%
where we replace the symmetric matrix variable $\mathbf S$ with its vector form that consists of $n \Def N(N-1)/2$ unique perturbation  variables in $\mathbf S$. We recall that $f_i$ could be either a CE-type or a CW-type per-node attack loss. In the case b) of re-trainable $\mathbf W$, the attack generation problem has the following min-max form
{\small \begin{align}\label{eq:  robust_attack}
    \begin{array}{cc}
\displaystyle \minimize_{  \mathbf 1^\top \mathbf s \leq  \epsilon,   \mathbf s\in \{ 0,1\}^{n} }   \displaystyle \maximize_{\mathbf W}  \, \sum_{i \in \mathcal V}f_i(\mathbf s ,  \mathbf W ;  \mathbf A,  \{ \mathbf x_i\}, y_i )  ,
    \end{array}
\end{align}}%
where the inner maximization aims to constrain the attack loss by retraining $\mathbf W$ so that attacking GNN is more difficult.

Motivated by targeted adversarial attacks against image classifiers \cite{carlini2017towards}, 
we can
define targeted topology attacks that are restricted to perturb edges of  targeted nodes. In this case,  we  require to linearly constrain $\mathbf S$ in \eqref{eq:A_indci} as
 $S_{i, \cdot} = 0$ if $i$ is not a target node.
 As a result, both attack formulations \eqref{eq:  prob_attack_s} and \eqref{eq:  robust_attack} have   extra linear constraints with respect to $s$, which can be readily handled by the optimization solver introduced later. Without loss of generality, we focus on untargeted topology attacks in this paper.

\subsection{PGD Topology Attack}

Problem   \eqref{eq:  prob_attack_s} is a combinatorial optimization problem due to the presence of Boolean variables. For ease of optimization,  we  relax $\mathbf s \in \{ 0,1\}^{n}$ to its convex hull $s \in [ 0,1]^{n}$ and solve the resulting continuous optimization problem,
{\small \begin{align}\label{eq:  prob_attack_cont}
    \begin{array}{ll}
\displaystyle \minimize_{\mathbf s}         & f(\mathbf s) \Def  \sum_{i \in \mathcal V}f_i(\mathbf s  ; \mathbf W,  \mathbf A,  \{ \mathbf x_i\}, y_i ) \\
      \st    & \mathbf s \in \mathcal S,
    \end{array}
\end{align}}%
where $\mathcal S = \{ \mathbf s \, | \, \mathbf 1^T s \leq \epsilon, \mathbf s \in [0,1]^n\}$. Suppose that the solution of problem \eqref{eq:  prob_attack_cont} is achievable, the remaining question is how to recover a binary solution from it. Since the variable $\mathbf s$ in \eqref{eq:  prob_attack_cont} can be interpreted as a probabilistic vector,    a randomization
sampling \cite{liu2016sensor} is suited for generating a near-optimal binary topology perturbation; see details in Algorithm\,\ref{alg: random_sample}.

\begin{algorithm}
\caption{Random sampling from probabilistic to  binary topology perturbation}
\begin{algorithmic}[1]
\State Input: probabilistic vector $\mathbf s$, $K$ is \# of random trials
\For{$k =  1,2,\ldots, K$}
\State draw binary vector $\mathbf u^{(k)}$ following {\small \begin{align}
   u^{(k)}_i = \left \{ \begin{array}{ll}
    1 &  \text{with probability $s_{i}$}\\
    0 &  \text{with probability $1- s_{i}$}
\end{array} \right. , \forall i 
\end{align}}%
\EndFor  
\State choose a vector $\mathbf s^*$ from $\{ \mathbf u^{(k)} \}$ which yields the smallest
attack loss $f(\mathbf u^{(k)} )$ under $\mathbf 1^T \mathbf s \leq \epsilon $. 
\end{algorithmic}\label{alg: random_sample}
\end{algorithm}


We solve the continuous optimization problem \eqref{eq:  prob_attack_cont} by projected gradient descent (PGD),
{\small \begin{align}\label{eq: PGD_step}
 \mathbf s^{(t)} = \Pi_{\mathcal S} \left [ \mathbf s^{(t-1)} - \eta_{t} \hat{\mathbf g}_t  \right ],
\end{align}}%
where $t$ denotes the iteration index of PGD, $\eta_t > 0$ is the learning rate at iteration $t$,
$\hat{\mathbf g}_t = \nabla f(\mathbf s^{(t-1)})$ denotes the gradient of the attack loss $f$ evaluated at $\mathbf s^{(t-1)}$, 
and $\Pi_{\mathcal S} (\mathbf a) \Def \argmin_{\mathbf s \in \mathcal S} \| \mathbf s - \mathbf a \|_2^2$ is the projection operator at $\mathbf a$ over the constraint set $\mathcal S$. In  Proposition\,\ref{prop:    proj_l1}, we show that the projection operation yields the closed-form solution.

\begin{myprop}\label{prop:    proj_l1}
Given $\mathcal S = \{ \mathbf s \, | \, \mathbf 1^T s \leq \epsilon, \mathbf s \in [0,1]^n\}$, the projection operation at the point $\mathbf a$ with respect to $\mathcal S$
 is
{\small \begin{align}\label{eq:  proj_sol_v1}
    \Pi_{\mathcal S} (\mathbf a) =  \left \{ 
    \begin{array}{ll}
          P_{[\mathbf 0, \mathbf 1]} [\mathbf a - \mu \mathbf 1] &  
          \begin{array}{l}
               \text{If $\mu > 0$ and}  \\
               \mathbf 1^T  P_{[\mathbf 0, \mathbf 1]} [\mathbf a - \mu \mathbf 1] = \epsilon,
          \end{array} \\
          & \\
           P_{[\mathbf 0, \mathbf 1]} [\mathbf a ] & \text{If $\mathbf 1^T P_{[\mathbf 0, \mathbf 1]} [\mathbf a ]  \leq \epsilon$},
    \end{array}
    \right.
\end{align}}%
where $P_{[0,1]} (x ) = x$ if $x \in [0,1]$, $0$ if $x < 0$, and $1$ if $x > 1$.
\end{myprop}
\textbf{Proof}:
We express the projection problem   as
\begin{align}\label{eq:  prob_proj_v1}
    \begin{array}{ll}
        \displaystyle \minimize_{\mathbf s} & \frac{1}{2}\| \mathbf s - \mathbf a \|_2^2 + \mathcal I_{[\mathbf 0, \mathbf 1]}(\mathbf s)\\
        \st  & \mathbf 1^\top \mathbf s \leq \epsilon, 
    \end{array}
\end{align}
where $\mathcal I_{[\mathbf 0, \mathbf 1]}(\mathbf s) = 0$ if $\mathbf s \in [0,1]^{n}$, and $\infty$ otherwise.

The Lagrangian function of  problem \eqref{eq:  prob_proj_v1} is given by
{\small \begin{align}
    & \frac{1}{2} \| \mathbf s - \mathbf a \|_2^2 + \mathcal I_{[\mathbf 0, \mathbf 1]}(\mathbf s) + \mu (\mathbf 1^\top \mathbf s - \epsilon) \nonumber \\
    = & \sum_{i} \left ( \frac{1}{2} (s_i - a_i)^2 + \mathcal I_{[0,1]}(s_i) + \mu s_i \right ) - \mu \epsilon,
\end{align}}%
where $\mu \geq 0$ is the dual variable. 
The minimizer to the above Lagrangian function (with respect to the variable $\mathbf s$) is 
{\small \begin{align}\label{eq:  station_vec}
    \mathbf s = P_{[\mathbf 0, \mathbf 1]} (\mathbf a - \mu \mathbf 1 ), 
\end{align}}%
where $P_{[\mathbf 0, \mathbf 1]}$ is taken   elementwise.
Besides the stationary condition \eqref{eq:  station_vec}, 
 other KKT conditions for 
solving problem \eqref{eq:  prob_proj_v1} are  
{\small \begin{align}
&    \mu(\mathbf 1^\top \mathbf s - \epsilon) = 0 , \label{eq:  primal}\\
& \mu \geq 0, \\
& \mathbf 1^\top \mathbf s \leq \epsilon. \label{eq:  feasibility}
\end{align}}%
If $\mu > 0$, then the solution to problem \eqref{eq:  prob_proj_v1} is given by \eqref{eq:  station_vec}, where the dual variable $\mu $ is determined by \eqref{eq:  station_vec} and \eqref{eq:  primal}
{\small \begin{align}
   \mathbf 1^T  P_{[\mathbf 0, \mathbf 1]} [\mathbf a - \mu \mathbf 1] = \epsilon,
~\text{and}~   \mu > 0.
\end{align}}%
If $\mu = 0$, then the solution to problem \eqref{eq:  prob_proj_v1} is given by \eqref{eq:  station_vec} and \eqref{eq:  feasibility},
{\small \begin{align}
    \mathbf s = P_{[\mathbf 0, \mathbf 1]} (\mathbf a),~\text{and}~ \mathbf 1^\top \mathbf s \leq \epsilon,
\end{align}}%
The proof is complete. \hfill $\square $

In the projection operation \eqref{eq:  proj_sol_v1}, one might need to solve the scalar equation $\mathbf 1^T  P_{[\mathbf 0, \mathbf 1]} [\mathbf a - \mu \mathbf 1] = \epsilon$
with respect to the dual variable $\mu$. 
This can be accomplished by applying the bisection method \cite{boyd2004convex,liu2015sparsity} over $\mu \in [\min(\mathbf a-\mathbf 1), \max(\mathbf a) ]$. That is because 
$\mathbf 1^T  P_{[\mathbf 0, \mathbf 1]} [\mathbf a - \max(\mathbf a) \mathbf 1] \leq  \epsilon$ 
and 
$\mathbf 1^T  P_{[\mathbf 0, \mathbf 1]} [\mathbf a - \min(\mathbf a - \mathbf 1) \mathbf 1 ] \geq  \epsilon$, where $\max$ and $\min$ return the largest and smallest entry of a vector. We remark that the bisection method converges in the logarithmic rate given by
$\log_2 {[ (\max(\mathbf a)  -\min(\mathbf a-\mathbf 1) )/\xi]}$ for the solution of $\xi$-error tolerance. 
We summarize the PGD topology attack in Algorithm\,\ref{alg: pgd_GCN}.

\begin{algorithm}
\caption{PGD topology attack on GNN}
\begin{algorithmic}[1]
\State Input: $\mathbf s^{(0)}$, $\epsilon > 0$,  learning rate $\eta_t$, and  iterations $T$
\For{$t =  1,2,\ldots, T$}
\State  gradient descent: $\mathbf a^{(t)}  = \mathbf s^{(t-1)} - \eta_{t} \nabla f(\mathbf s^{(t-1)}) $
\State call projection operation in \eqref{eq:  proj_sol_v1}
\EndFor  
\State call Algorithm\,\ref{alg: random_sample} to return $\mathbf s^*$, and the resulting $\mathbf A^\prime$ in \eqref{eq:A_indci}.
\end{algorithmic}\label{alg: pgd_GCN}
\end{algorithm}

\subsection{Min-max Topology Attack}
We next solve the problem of min-max attack generation in \eqref{eq:  robust_attack}. By convex relaxation on the Boolean variables, we obtain the following continuous optimization problem
{\small \begin{align}\label{eq:  robust_attack_cont}
 \hspace*{-0.15in}\begin{array}{cc}
\displaystyle \minimize_{   \mathbf s\in \mathcal S }   \displaystyle \maximize_{\mathbf W}  \, f(\mathbf s, \mathbf W) = \sum_{i \in \mathcal V}f_i(\mathbf s ,  \mathbf W ;  \mathbf A,  \{ \mathbf x_i\}, y_i )  ,
    \end{array}
\end{align}}%
where $\mathcal S$ has been defined in \eqref{eq:  prob_attack_cont}. We solve problem  \eqref{eq:  robust_attack_cont} by first-order alternating optimization \cite{lu2019understand,lutsahong18}, where the inner maximization is solved by gradient ascent, and the outer minimization is handled by PGD same as \eqref{eq: PGD_step}.  We summarize the min-max topology attack in Algorithm\,\ref{alg: min_max_attack}.
We remark that one can  perform multiple maximization  steps  within each  iteration  of alternating optimization.  This strikes a balance between the computation efficiency and the convergence accuracy \cite{chen2017robust,Qian2018RobustOO}. 

\begin{algorithm}
\caption{Min-max topology attack to solve \eqref{eq:  robust_attack_cont}}
\begin{algorithmic}[1]
\State Input: given $\mathbf W^{(0)}$, $\mathbf s^{(0)}$, learning rates $\beta_t$ and $\eta_t$, and iteration numbers $T$
\For{$t =  1,2,\ldots, T$}
\State inner maximization over $\mathbf W$: given  $\mathbf s^{(t-1)}$, obtain  
{\small \begin{align*}
    \mathbf W^{t} = \mathbf W^{t-1} + \beta_t \nabla_{\mathbf W} f (\mathbf s^{t-1},\mathbf W^{t-1})
\end{align*}}%
\State outer minimization over $\mathbf s$: given  $\mathbf W^{(t)}$,   running \hspace*{0.18in}
PGD \eqref{eq: PGD_step}, where $\hat{\mathbf g}_t = \nabla_{\mathbf s} f (\mathbf s^{t-1},\mathbf W^{t})$
\EndFor  
\State call Algorithm\,\ref{alg: random_sample} to return $\mathbf s^*$, and the resulting $\mathbf A^\prime$ in \eqref{eq:A_indci}.
\end{algorithmic}\label{alg: min_max_attack}
\end{algorithm}

\section{Robust Training for GNNs}

With the aid of first-order attack generation methods, we now introduce our adversarial training for GNNs via robust optimization. Similar formulation is also used in \cite{madry2017towards}. In adversarial training, we solve a min-max problem for robust optimization:
{\small \begin{align}\label{eq:  robust_train}
    \begin{array}{cc}
\displaystyle \minimize_{\mathbf W}   \maximize_{ \mathbf s \in \mathcal S} \,  -f(\mathbf s, \mathbf W) ,
    \end{array}
\end{align}}%
where $ f(\mathbf x, \mathbf W)$ denotes the attack loss specified in \eqref{eq:  robust_attack_cont}. Following the idea of adversarial training for image classifiers in \cite{madry2017towards}, we restrict the loss function $f$ as the CE-type loss. This formulation tries to minimize the training   loss at the presence of topology  perturbations. 

We note that problems  \eqref{eq:  robust_train} and \eqref{eq:  robust_attack} share a very similar min-max form, however, they are not equivalent since the loss $f$ is neither convex with respect to $\mathbf s$ nor concave with respect to $\mathbf W$, namely, lacking   saddle point property~\cite{boyd2004convex}. However, there exists connection   between \eqref{eq:  robust_attack} and \eqref{eq:  robust_train}; see  Proposition\,\ref{prop: min_max_connection}. 

\begin{myprop}\label{prop: min_max_connection}
Given a general attack loss function $f$, problem  \eqref{eq:  robust_train} is equivalent to
{\small \begin{align}\label{eq: min_max_sol_same}
    \maximize_{\mathbf W}   \minimize_{ \mathbf s \in \mathcal S} \,   f(\mathbf s, \mathbf W),
\end{align}}%
which further yields $\eqref{eq: min_max_sol_same} \leq \eqref{eq:  robust_attack_cont}$.
\end{myprop}
\textbf{Proof}: By introducing epigraph variable  $p$ \cite{boyd2004convex}, problem \eqref{eq:  robust_train} can be rewritten as
{\small \begin{align}\label{eq: prob_minmax_1}
    \begin{array}{ll}
       \displaystyle \minimize_{\mathbf W, p}     &  p \\
    \st      & -
    f(\mathbf s, \mathbf W) \leq p, \forall \mathbf s \in \mathcal S.
    \end{array}
\end{align}}%
By changing variable $q \Def - p$, problem \eqref{eq: prob_minmax_1} is equivalent to
{\small \begin{align}\label{eq: prob_minmax_2}
    \begin{array}{ll}
       \displaystyle \maximize_{\mathbf W, q}     &  q \\
    \st      &  
    f(\mathbf s, \mathbf W) \geq q, \forall \mathbf s \in \mathcal S.
    \end{array}
\end{align}}%
By eliminating the epigraph variable $q$, 
problem \eqref{eq: prob_minmax_2} becomes \eqref{eq: min_max_sol_same}. By \textit{max-min inequality}~\cite[Sec.\,5.4]{boyd2004convex}, we finally obtain that
{\small
\begin{align*}
    \maximize_{\mathbf W}   \minimize_{ \mathbf s \in \mathcal S} \,   f(\mathbf s, \mathbf W) \leq   \minimize_{   \mathbf s\in \mathcal S }    \maximize_{\mathbf W}  \, f(\mathbf s, \mathbf W).
\end{align*}
}%
The proof is now complete. \hfill $\square$
%

We summarize the robust training algorithm in Algorithm\,\ref{alg: adv_training} for solving problem 
\eqref{eq: min_max_sol_same}.
Similar to Algorithm\,\ref{alg: min_max_attack}, one usually performs multiple inner minimization  steps (with respect to $\mathbf s$) within each  iteration $t$ to have a solution towards minimizer during alternating optimization. This  improves the stability of convergence in practice \cite{Qian2018RobustOO,madry2017towards}.

\begin{algorithm}
\caption{Robust training for solving problem \eqref{eq: min_max_sol_same}}
\begin{algorithmic}[1]
\State Input: given $\mathbf W^{(0)}$, $\mathbf s^{(0)}$, learning rates $\beta_t$ and $\eta_t$, and iteration numbers $T$
\For{$t =  1,2,\ldots, T$}
\State inner minimization over $\mathbf s$: given  $\mathbf W^{(t-1)}$,   running \hspace*{0.18in}
PGD \eqref{eq: PGD_step}, where $\hat{\mathbf g}_t = \nabla_{\mathbf s} f (\mathbf s^{t-1},\mathbf W^{t-1})$
\State outer maximization over $\mathbf W$: given  $\mathbf s^{(t)}$, obtain  
{\small \begin{align*}
    \mathbf W^{t} = \mathbf W^{t-1} + \beta_t \nabla_{\mathbf W} f (\mathbf s^{t},\mathbf W^{t-1})
\end{align*}}%
\EndFor  
\State return $\mathbf W^{T}$.
\end{algorithmic}\label{alg: adv_training}
\end{algorithm}

\section{Experiments}

In this section, we present our experimental results for both topology attack and defense methods on  a graph convolutional networks (GCN) \cite{kipf2016semi}. We demonstrate the misclassification rate and the convergence of the   proposed $4$ attack methods: negative cross-entropy loss via PGD attack (CE-PGD), CW loss via PGD attack (CW-PGD), negative cross-entropy loss via min-max attack (CE-min-max), CW loss via min-max attack (CW-min-max). We then show the improved robustness of  GCN by leveraging our proposed  robust training against  topology attacks.   

\subsection{Experimental Setup}
We evaluate our methods on two well-known datasets: Cora and Citeseer \cite{sen2008collective}. Both datasets contain unweighted edges which can be generated as symmetric adjacency matrix $\mathbf A $ and sparse bag-of-words feature vectors which can be treated the input  of GCN. 
To train the model, all node feature vectors  are fed into GCN but with only 
  140 and 120 labeled nodes for Cora and Citeseer, respectively. 
  The number of test labeled nodes is 1000 for both datasets. 
At each experiment, we repeat $5$ times based on  different splits of training/testing nodes and report \textit{mean} $\pm$ \textit{standard deviation} of misclassification rate (namely, 1 $-$ prediction accuracy) on testing nodes. 

\subsection{Attack Performance}
We compare our four attack methods (CE-PGD, CW-PGD, CE-min-max, CW-min-max) with DICE (`delete edges internally, connect externally’) ~\cite{waniek2018hiding}, Meta-Self attack \cite{zugner2019adversarial} and greedy attack, a variant of Meta-Self attack   without  weight re-training for GCN. The greedy attack is considered as a fair comparison with our 
 CE-PGD and CW-PGD attacks, which  are generated on a fixed GCN without weight re-training. 
 In min-max attacks (CE-min-max and CW-min-max), we show   misclassification rates against both natural   and retrained models from Algorithm\,\ref{alg: min_max_attack}, and compare them with the state-of-the-art Meta-Self attack. 
 For a fair comparison, we use the same performance evaluation criterion in Meta-Self, testing nodes' predicted labels (not their ground-truth label) by an independent pre-trained model that can be used during the attack. 
In the attack problems \eqref{eq:  prob_attack_s} and \eqref{eq:  robust_attack},  unless specified otherwise
 the maximum number of perturbed edges  is set to be $5\%$ of the total number of existing edges in the original graph.
In   Algorithm \ref{alg: random_sample}, we set the iteration number of random sampling as $K = 20$
and choose the perturbed topology  with the highest misclassification rate which also satisfies the edge perturbation constraint. 

In Table\,\ref{table_attack_performance},
we present the misclassification rate of different attack methods against both natural and retrained model from \eqref{eq:  robust_attack_cont}. Here we recall that the retrained model arises due to the scenario of attacking an interactive GCN with re-trainable weights (Algorithm\,\ref{alg: min_max_attack}).
For comparison, we also show the misclassification  rate of a natural model with the true topology (denoted by `clean'). As we can see, to attack the natural model, our proposed attacks achieve better misclassification  rate  than the existing methods. We also observe that compared to  min-max attacks (CE-min-max and CW-min-max), CE-PGD and CW-PGD yield better attacking performance since it is easier to attack a pre-defined GCN. To attack the   model that  allows retraining, we set $20$ steps  of inner maximization per iteration of Algorithm\,\ref{alg: min_max_attack}. The results show that our proposed min-max attack  achieves very competitive performance compared to Meta-Self attack. Note that evaluating the attack performance on the retrained model obtained from \eqref{eq:  robust_attack_cont} is not quite fair since the retrained weights could be sub-optimal and   induce degradation in  classification.


\begin{table}[htb]
 \centering
\begin{tabular}{c|c|cc}
\toprule[1pt]
\multirow{7}{*}{\begin{tabular}[c]{@{}c@{}}\\ \\fixed\\natural\\ model\end{tabular}} 
          && Cora & Citeseer \\
\cline{2-4}
&clean     & $18.2 \pm 0.1$ & $28.9 \pm 0.3 $    \\  
\cline{2-4}
&DICE      & $18.9\pm 0.2$ & $29.8\pm 0.4$     \\

&Greedy    & $25.2\pm 0.2$ & $34.6\pm 0.3$     \\
&Meta-Self & $22.7\pm 0.3$ & $31.2\pm 0.5$        \\
\cline{2-4}
&\textbf{CE-PGD}    & $\mathbf{28.0\pm 0.1}$ & $36.0\pm 0.2$     \\
&\textbf{CW-PGD}    & $27.8\pm 0.4$ & $\mathbf{37.1\pm 0.5}$     \\
&\textbf{CE-min-max} & $26.4\pm 0.1$  &  $34.1\pm 0.3$        \\
& \textbf{CW-min-max} & $26.0\pm 0.3$  &  $34.7\pm 0.6$       \\

\midrule[1pt]
\multirow{3}{*}{\begin{tabular}[c]{@{}c@{}}retrained\\ model \\from \eqref{eq:  robust_attack_cont}\end{tabular}} 
&Meta-Self &  $29.6\pm 0.4$    &    $\mathbf{39.7\pm 0.3}$       \\
\cline{2-4}
& \textbf{CE-min-max} &  $\mathbf{30.8\pm 0.2}$    & $37.5\pm 0.3$         \\
& \textbf{CW-min-max} &  $30.5\pm 0.5$    & $39.6\pm 0.4$       \\

\bottomrule[1pt]

\end{tabular}
 \caption{Misclassification rates ($\%$) under $5\%$ perturbed edges
  } 
   \label{table_attack_performance}
\end{table}

\begin{figure}[htb]  
\centering
\vspace*{-0.15in}
\hspace*{-0.15in}
\begin{tabular}{cc}
 \includegraphics[width=.24\textwidth]{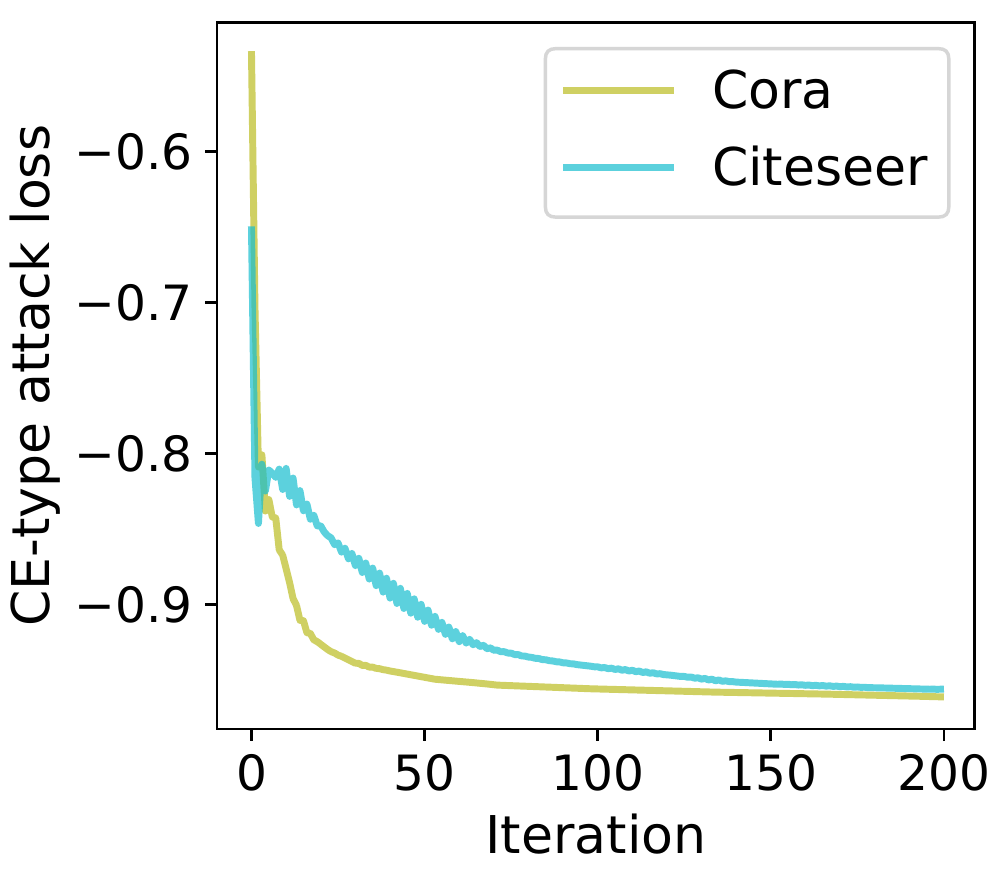} & \hspace*{-0.15in}
 \includegraphics[width=.24\textwidth]{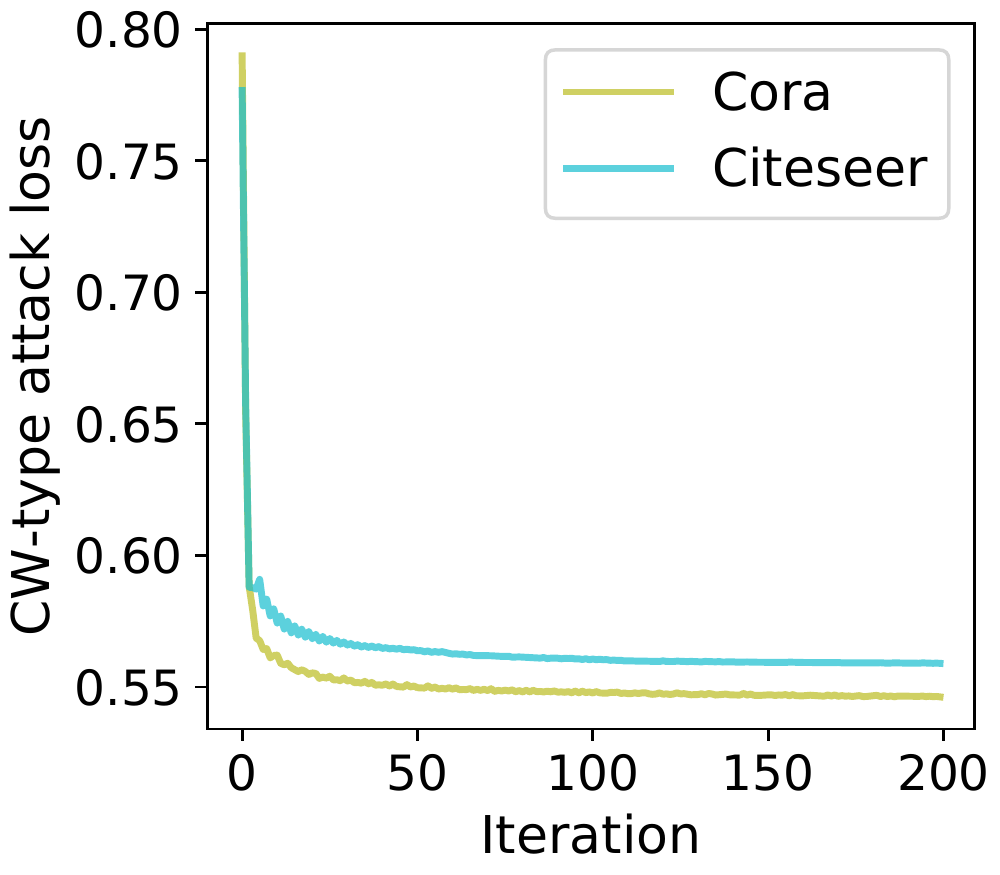} 
\vspace*{-0.15in}
 \end{tabular}
\caption {CE-PGD and CW-PGD attack losses on Cora and Citeseer datasets.
}
\label{fig: attack_loss}
\end{figure}

In Fig.\,\ref{fig: attack_loss}, we present the CE-loss and the CW-loss of the proposed topology attacks against the number of iterations in Algorithm\,\ref{alg: pgd_GCN}. Here we choose $T = 200$ and $\eta_t = 200/\sqrt{t}$. As we can see, the method of PGD converges gracefully against iterations. This verifies the effectiveness of the   first-order  optimization based attack  generation method. 

\vspace*{-0.15in}

\subsection{Defense Performance}

In what follows, we   invoke Algorithm\,\ref{alg: adv_training} to generate robust GCN via adversarial training.  
 We set $T = 1000$, $\beta_t = 0.01$ and $\eta_t = 200/\sqrt{t}$. We run $20$ steps for inner minimization. Inspired by \cite{madry2017towards}, we increase the hidden units from 16 to 32 in order to create more capacity for this more complicated classifier.
Initially, we set the maximum number of edges we can modify as $5\%$ of total existing edges. 

\begin{figure}[htb]  
\centering
\vspace*{-0.1in}
\begin{tabular}{c}
 \includegraphics[width=.43\textwidth]{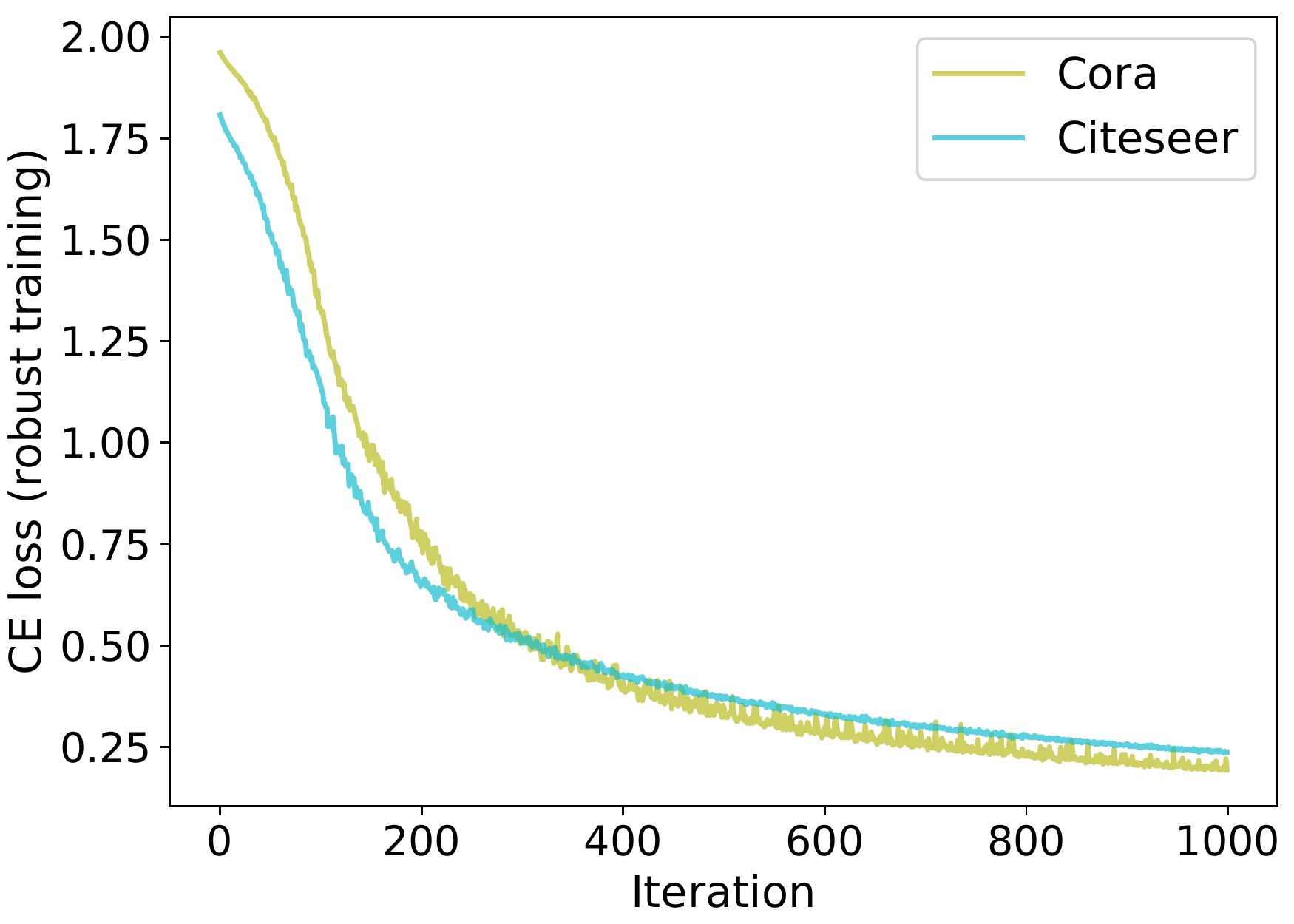}
\vspace*{-0.15in}
 \end{tabular}
\caption {Robust training loss on Cora and Citeseer datasets. }
\label{fig: robust_training_loss}
\vspace*{-0.15in}
\end{figure}

\begin{figure}[htb]  
\centering
\hspace*{-0.05in}
\begin{tabular}{c}
 \includegraphics[width=.43\textwidth]{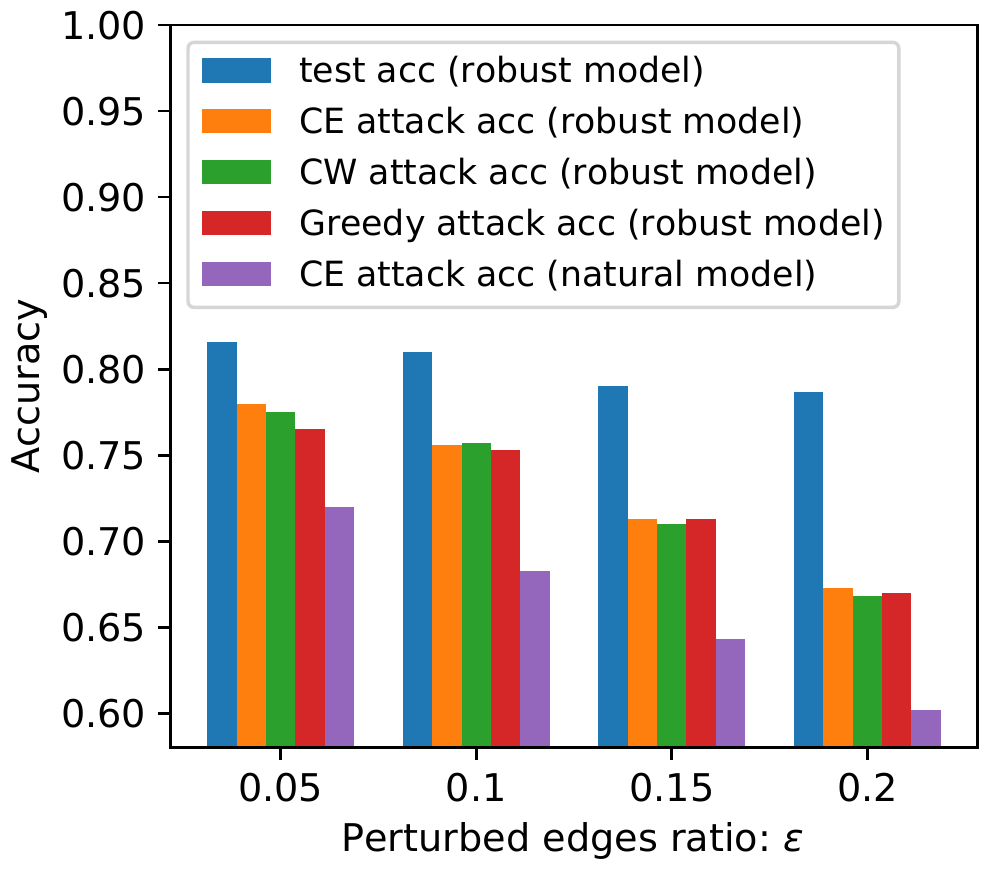}
\vspace*{-0.15in}
 \end{tabular}
\caption {Test accuracy of robust model (no attack), 
CE-PGD attack against robust model,   CW-PGD attack against robust model, Greedy attack against robust model and CE-PGD attack against natural model for different $\epsilon$ used in robust training and test on Cora dataset. }
\label{fig: adv_comparision}
\vspace*{-0.15in}
\end{figure}

In Figure\,\ref{fig: robust_training_loss}, we present convergence of our robust training. As we can see, the loss drops reasonably and the $1,000$ iterations are necessary for robust training rather than normal training process which only need $200$ iterations. We also observe that our robust training algorithm does not harm the test accuracy when $\epsilon = 5\%$, but successfully improves the robustness as the attack success rate drops from $28.0\%$ to $22.0\%$ in Cora dataset as shown in Table\,\ref{table_advtrain_performance}, 

After showing the effectiveness of our algorithm, we explore deeper in adversarial training on GCN. We aim to show how large $\epsilon$ we can use in robust training. So we set $\epsilon$ from $5\%$ to $20\%$ and apply CE-PGD attack following the same $\epsilon$ setting. The results are presented in Table\,\ref{table_advtrain_matrix}. Note that when $\epsilon=0$, the first row shows  misclassification rates of test nodes on natural graph as the baseline for \textit{lowest} misclassification rate we can obtain; the first column shows the CE-PGD attack misclassification rates of natural model as the baseline for \textit{highest} misclassification rate we can obtain. 
We can conclude that when a robust model trained under an $\epsilon$ constraint, the model will gain robustness under this  $\epsilon$ distinctly. Considering its importance to keep the original graph test performance, we suggest generating robust model under  $\epsilon = 0.1$. Moreover, please refer to Figure\,\ref{fig: adv_comparision} that a) our robust trained model can provide universal defense to CE-PGD, CW-PGD and Greedy attacks; b) when increasing $\epsilon$, the difference between both test accuracy and CE-PGD attack accuracy increases substantially, which also implies the robust model under larger $\epsilon$ is harder to obtain.

\begin{table}[htb]
 \centering
\begin{tabular}{c|cc}
\toprule[1pt]
                  & Cora & Citeseer \\
\midrule[1pt]
 $\mathbf A  $/natural model    & $18.2\pm 0.1$ & $28.9\pm 0.1$     \\
 $\mathbf A  $/robust model & $18.1\pm 0.3$ & $28.7\pm 0.4$     \\
 $\mathbf A^\prime  $/natural model    & $28.0\pm 0.1$ & $36.0\pm 0.2$     \\
 $\mathbf A^\prime  $/robust model & $22.0\pm 0.2$ & $32.2\pm 0.4$   \\
 
 \bottomrule[1pt]
\end{tabular}
 \caption{Misclassification rates ($\%$) of robust training (smaller is better for defense task) with at most $5\%$ of edge perturbations. $\mathbf A  $ means the natural graph, $\mathbf A^\prime  $ means the generated adversarial graph under $\epsilon = 5\%$. $\mathbf{X} / M$ means the misclassification rate of using model $M$ on graph $\mathbf{X}$. 
  } 
  \label{table_advtrain_performance}
\end{table}

\begin{table}[htb]
  \centering
\begin{tabular}{c|cccccc}
\toprule[1pt]
                            & \multicolumn{6}{c}{$\epsilon$ in robust training (in \%)} \\ \hline
\multirow{6}{*}{\begin{tabular}[c]{@{}c@{}}$\epsilon$ in\\ attack \\(in \%)\end{tabular}}
                            &      &  0    & 5        & 10     & 15     & 20     \\
                            & 0    &$18.1$ & $18.2$   & $19.0$   & $20.2$   & $21.3$   \\
                            & 5    &$27.9$ & $22.0$   & $23.9$   & $24.8$   & $26.5$   \\
                            & 10   &$32.7$ & $32.1$   & $26.4$   & $27.7$   & $31.0$    \\
                            & 15   &$36.7$ & $36.2$   & $33.4$   & $29.7$   & $32.9$   \\
                            & 20   &$40.2$ & $40.1$   & $36.3$   & $36.3$   & $33.5$  \\
 \bottomrule[1pt]
\end{tabular}
 \caption{Misclassification rates ($\%$) of CE-PGD attack against robust training model versus   (smaller is better)  different $\epsilon$  ($\%$)   on Cora dataset. Here
 $\epsilon=0$ in training means natural model and $\epsilon=0$ in attack means unperturbed topology.
  }
     \label{table_advtrain_matrix}
\end{table}

\vspace{-0.5cm}
\section{Conclusion}
In  this  paper,  we first introduce an edge perturbation based topology attack framework that  overcomes the difficulty of attacking discrete graph structure data from a first-order optimization perspective. Our extensive experiments show that with only a fraction of edges changed, we are able to compromise state-of-the-art graph neural networks model noticeably. Additionally, we propose an adversarial training framework to improve the robustness of GNN models based on our attack methods. Experiments on different datasets show that our method is able to improve the GNN model's robustness against both gradient based and greedy search based attack methods without classification performance drop on original graph. We believe that this paper provides potential means for theoretical study and improvement of the robustness of deep learning models on graph data.

\section*{Acknowledgments}

This work is supported by Air Force Research Laboratory FA8750-18-2-0058
and the MIT-IBM Watson AI Lab.

\clearpage

\bibliographystyle{named}
\bibliography{ijcai19}




\end{document}